\documentclass{article}

\usepackage{arxiv}

\usepackage[utf8]{inputenc} %
\usepackage[T1]{fontenc}    %
\usepackage{hyperref}       %
\usepackage{url}            %
\usepackage{booktabs}       %
\usepackage{amsfonts}       %
\usepackage{nicefrac}       %
\usepackage{microtype}      %
\usepackage{lipsum}

\usepackage[superscript,biblabel]{cite}

\hypersetup{colorlinks}

\usepackage{tikz}
\usepackage{tikzscale}
\usetikzlibrary{arrows}
\usepackage{tikzscale}
\usepackage{textcomp}
\usepackage{soul,color}

\usepackage{afterpage}
\usepackage{tabularx}
\usepackage{tabularx}
\usepackage{booktabs}
\usepackage{colortbl}
\usepackage{xcolor}
\usepackage{xfrac}
\usepackage{url}

\newcommand{\cpp}{{C\nolinebreak[4]\hspace{-.05em}\raisebox{.4ex}{\tiny\bf ++}}}

\newcommand{\apis}{\textsc{api}s}

\title{Gym-Ignition: Reproducible Robotic Simulations for Reinforcement Learning}

\author{
  Diego Ferigo\thanks{Dynamic Interaction Control, Istituto Italiano di Tecnologia} ${}^{\ ,}$ \thanks{University of Manchester, Machine Learning and Optimisation} \\
  \texttt{diego.ferigo@iit.it} \\
  \And
  Silvio Traversaro ${}^{*}$ \\
  \texttt{silvio.traversaro@iit.it} \\
  \And
  Giorgio Metta ${}^{*}$ \\
  \texttt{giorgio.metta@iit.it}
  \And
  Daniele Pucci ${}^{*}$\\
  \texttt{daniele.pucci@iit.it}
}

\date{}

\begin{document}
\maketitle

\begin{abstract}
This paper presents Gym-Ignition, a new framework to create reproducible robotic environments for reinforcement learning research.
It interfaces with the new generation of Gazebo, part of the Ignition Robotics suite, which provides three main improvements for reinforcement learning applications compared to the alternatives: 1) the modular architecture enables using the simulator as a \cpp{} library, simplifying the interconnection with external software; 2) multiple physics and rendering engines are supported as plugins, simplifying their selection during the execution; 3) the new distributed simulation capability allows simulating complex scenarios while sharing the load on multiple workers and machines.
The core of Gym-Ignition is a component that contains the Ignition Gazebo simulator and exposes a simple interface for its configuration and execution. We provide a Python package that allows developers to create robotic environments simulated in Ignition Gazebo. Environments  expose the common OpenAI Gym interface, making them compatible out-of-the-box with third-party frameworks containing reinforcement learning algorithms. Simulations can be executed in both headless and GUI mode, the physics engine can run in accelerated mode, and instances can be parallelized. Furthermore, the Gym-Ignition software architecture provides abstraction of the Robot and the Task,
making environments agnostic on the specific runtime. This abstraction allows their execution also in a real-time setting on actual robotic platforms, even if driven by different middlewares.
\end{abstract}


\section{Introduction}
Simulation has always been a key component in robotics.
Over the years, its accuracy and efficiency constantly improved, and nowadays there are numerous valid physics engines and simulators.
It became part of every roboticist's toolbox and always collected great interest and contributions from the entire community.
In particular, in order to solve their decision making problems, agents trained with Reinforcement Learning (RL) algorithms need to dynamically interact with an environment by taking actions and gathering information of their consequences --- i.e. sampling experience.
Classical benchmarks used in this research field typically involve grid-worlds or simple toy problems.
The advent of Deep Learning (DL) and its combination with RL, however, enabled machines to solve complex decision making tasks that have been out of their reach until now.

New benchmarks involving harder and more complex scenarios --- environments --- have been developed, mainly originating from the gaming realm.
A virtuous cycle we experienced in this domain is the constant push of novel algorithms thanks to more complex environments and vice versa. The community has been very prolific in constantly extending the offer of these environments. Typical examples are the Arcade Learning Environment~\cite{bellemareArcadeLearningEnvironment2013s}, OpenAI Gym~\cite{brockmanOpenAIGym2016s}, DeepMind Lab~\cite{beattieDeepMindLab2016s}, and DeepMind Control Suite~\cite{tassaDeepMindControlSuite2018s}, to name just a few. 

The new boosted capability of Deep RL attracted much interests from many research fields.
Robotics is one of those, and can benefit the most from
the freedom framed by the formulation of the Reinforcement Learning problems.
Applications range from manipulation to locomotion, both affected by a very high complexity which generally demands tedious heuristic tuning.
However, the interaction with the real world poses a set of challenges that differs considerably from the typical experimental setup of reinforcement learning, that involves gaming-like simulations. In a recent study~\cite{dulac-arnoldChallengesRealWorldReinforcement2019s}, authors outline nine high-level challenges of applying RL to the real world. Robotics suffers from all of them.

Analogously to classic robotic research, also the application of RL to robotics has always tried to take advantage of simulated environments.
The motivations, in this case, are even more critical since the systems to control are costly and delicate.
The intrinsic need for exploration during the training phase might be dangerous to the robotic platforms and their surroundings.
Moreover, even if safe constraints are enforced during training, collecting experience only using real-world interactions is often not sufficient.
Simulations can generate a large amount of synthetic experience that can be used to train policies. On one hand, simulations help overcoming the limitations of real-time data collection, but on the other they introduce a bias caused by intrinsic and unavoidable modeling approximations. The process of training a policy in simulation and then transferring it to the real world is better known as \emph{sim-to-real}~\cite{christianoTransferSimulationReal2016s, muratoreAssessingTransferabilitySimulation2019s}, and the difference between simulation and reality is typically addressed as \emph{reality gap}.

In the recent literature, many are the examples of successful attempts to port simulated policies to the real world. Common techniques to bypass the reality gap include improving the description of the simulated robot~
\cite{tanSimtoRealLearningAgile2018s}, learning effects difficult to characterize from real-world data and then using their models in simulation~\cite{chebotarClosingSimtoRealLoop2018s,hwangboLearningAgileDynamic2019s,jeongModellingGeneralizedForces2019s}, massively randomizing simulation parameters
\cite{pengSimtoRealTransferRobotic2018s, openaiLearningDexterousInHand2018s}, imitating behaviour from experts or existing controllers~\cite{liUsingDeepReinforcement2019s, xieIterativeReinforcementLearning2019s}, and applying hierarchical architecture to decompose complex tasks~\cite{jainHierarchicalReinforcementLearning2019s}.

A common denominator of all these works is the extensive use of complex simulations, most of the time using open-source software. In many cases, however, authors did not release their experimental setup, making the reproduction of their result very difficult if not impossible.
Reproducibility can be improved following two directions.
Firstly, the entire community would benefit from a standardized framework to apply reinforcement learning techniques to simulated robots. We believe that also in the robotics domain, standardized environments would trigger the same virtuous cycle that characterized the past breakthrough in reinforcement learning.
Secondly, the community would benefit from a platform that is versatile enough to minimize the system integration effort always required when dealing with real robotic platforms.
Roboticists know that real-world applicability involves a considerable amount of custom code and heuristic tuning. Though, simulating frameworks might at least try to abstract as much as possible low-level details and provide generic interfaces that can then be customized as needed. 

This work presents Gym-Ignition\footnote{\url{https://github.com/robotology/gym-ignition}}, a framework to create reproducible reinforcement learning environments for robotic research.
The environments created with Gym-Ignition target both simulated and real-time settings.
The modular project architecture enables to seamlessly switch between the two domains without the need to adapt the logic of the decision-making task to the real platform.
Simulated environments run in the new generation of the Gazebo simulator~\cite{koenigDesignUseParadigms2004s} called Ignition Gazebo and part of the Ignition Robotics\footnote{\url{https://ignitionrobotics.org}} suite. The simulator features a new abstraction layer that makes easy to integrate new physics engines in \cpp{} and switch between them during the simulation. Alternatively, new physics engines that provide Python bindings can be integrated at the Python level exploiting the Gym-Ignition abstraction layers.

Gym-Ignition aims to narrow the gap between reinforcement learning and robotic research.
It permits roboticists to create simulated environments using familiar tools like Gazebo, SDF\footnote{\url{http://sdformat.org/}}, and URDF\footnote{\url{http://wiki.ros.org/urdf}} files to model both the robot and the scenario.
Environment developers can choose either Python or \cpp, depending on the stage of the development. In fact, developers can benefit from the former for quick prototyping and the latter for deployment. For both domains, we provide the support of exposing the common OpenAI Gym interface~\cite{brockmanOpenAIGym2016s}. This makes the environments compatible with the majority of projects developed by the reinforcement learning community that provide algorithms.

To the best of our knowledge, Gym-Ignition is the first project that integrates with the new Ignition Robotics suite developed by Open Robotics\footnote{\url{https://www.openrobotics.org}}. We believe that it will progressively take over the current generation of the simulator, providing new features, enhanced performance, and improved user experience.

This paper is structured as follows. Firstly, we identify a set of useful properties that characterize projects providing robotic environments. Then, we selected the main available projects that provide robotic environments compatible with OpenAI Gym and briefly describe their properties and shortcomings. We proceed presenting Gym-Ignition, explaining its architecture and describing its features. Finally, we conclude discussing the current limitations and outlining the future activities.

\section{Background}

Nowadays, the variety of open- and closed-source simulation tools for rigid-body dynamics is huge and diverse in nature. Depending on the target application (gaming, robotics, etc.), existing tools might sacrifice physical accuracy for stability and performance. Defining a fair comparison setup goes beyond the scope of this paper, and for a detailed analysis of different physics engines we redirect the readers to specific works~\cite{ivaldiToolsSimulatingHumanoid2014s},~\cite{ erezSimulationToolsModelbased2015s}.

In this section, instead, we focus on defining high-level properties that we believe to be important for RL research applied to the robotics domain. To some extent, our analysis represents an extension of what described in~\cite{julianiUnityGeneralPlatform2018s}.

\textbf{Reproducibility} - In order to get reproducible simulations, different executions of the same experiment should yield the same outcome. A first common component that might compromise reproducibility is the presence of random number generators. In order to achieve reproducible results, simulation frameworks should expose to the users an interface to optionally initialize all the generator seeds. A second component that undermines reproducibility is a subtle consequence of the client-server architecture widely used by many simulators. Often, the physics engine and the user code that reads and writes values reside on different threads or processes. The communication between them relies on sockets whose processes, depending on the load of the system and scheduler of the operating system, can be preempted. In this way, in absence of complex synchronization protocols~\cite{krammerStandardizedIntegrationRealTime2019s}, the user code might think to have stepped the simulator and read the most recent measurement even if the data might have been buffered. In order to overcome this limitation, the simplest solution is to avoid as much as possible socket-based protocols.

\textbf{Modularity} - Reinforcement learning is one of the most generic learning frameworks. Its definition, a learning Agent that interacts with an Environment, facilitates the separation of the components in a Reinforcement Learning framework. In a sense, most of the learning-from-interaction processes can be represented by such scheme. In the field of robotics, however, we might benefit from a more fine-grained abstraction of the environment. For instance, we might want to achieve the same learning goal on robots with different mechanical structures.
In order to promote code reuse and minimize system integration effort, a modular design that abstracts the robot and the logic of the learned task is a valuable addition.

\textbf{Real-time Compatibility} - The main reason to rely on simulations for training an agent is the low cost of synthetic samples. However, the final goal should be the deployment of the policy on the real robotic platform. Frameworks should allow to either execute or continue the learning process on the real robot with minimal changes. An open problem, though, is how to reset a real-time rollout. For instance, in the case of floating-base robots, this operation may demand moving the robot back to the initial position in the operating area.

\textbf{Parallel Simulation} - Modern computers are nowadays endowed with multiple computational cores. The independence between simulated instances makes executing parallel environments trivial, maximizing local computing resources. On a higher level, the same applies when scaling to multiple machines. Typically, job distribution is performed by frameworks that provide the algorithms. Environment providers should only be careful of ensuring instances independence for multithread and multiprocess execution.

\textbf{Accelerated Simulation} - Simulations, based on their complexity, can run either faster or slower than real-time. The ratio between and real and simulated time is known as Real-Time Factor (RTF). A RTF greater then one indicates that the simulation is running in an accelerated state, i.e. faster than real-time. In order to speed up experience collection, environments should be able to run in accelerated mode.

\textbf{Headless Simulation} - A simulation is headless if it can be executed on a machine without any display. The definition of the learning goal is typically expressed in a mathematical form, therefore during the training process the visualization of the environment is not strictly necessary. Though, rendering the environment is helpful for fine-tuning the logic of the learning task and for debugging purposes.

\textbf{Multiple Physics Engines} - The physics engine is that simulation component which integrates physical equations, evaluates collisions, and solves contact constraints. Classic techniques in domain randomization~\cite{pengSimtoRealTransferRobotic2018s, ramosBayesSimAdaptiveDomain2019s} operate on parameters of the physics engine. Supporting multiple back-ends and being able to switch between them on-the-fly would permit the randomization of the entire physics engine, bringing domain randomization to a higher level while preventing that the learning agent overfits possible subtleties of a single implementation.

\textbf{Photorealistic Rendering} - Visuo-motor control is one of the main research directions in the field of reinforcement learning applied to robotics. The need for photorealistic rendering is a key component in this use case. Modern GPUs are becoming more capable of efficiently computing extremely complex light interaction in the simulated environment, and technologies such as ray tracing are becoming suitable for real-time usage.

\section{A Short Survey of Software for RL}

This section reports few projects that provide robotic environments compatible with OpenAI Gym and briefly describe their properties and shortcomings.

\textbf{OpenAI Robotic Environments}\footnote{\url{https://openai.com/blog/ingredients-for-robotics-research}} are part of the official OpenAI environments, which became the de-facto standard solution commonly used to benchmark algorithms. They are simulated with the Mujoco simulator~\cite{todorovMuJoCoPhysicsEngine2012s}, which became one of the most common solutions for continuous control tasks. Unfortunately, the simulator is proprietary software, constraint that greatly limits its use. Furthermore, the simulator has realistic rendering limitations.

\textbf{PyBullet Environments}~\cite{coumansPybulletPythonModule2016s} are part of the Bullet3 project and use Bullet as physics engine. Given the active development and open-source nature of the project, a big community circles around this physics engine. Simulations are reliable and fast, but the default rendering capabilities are not photorealistic. The provided robotic environments are valid, even if their documentation and modularity can be improved. PyBullet has an experimental Nvidia PhysX back-end (CPU based) and a real-time back-end might be open-sourced in the near future.

\textbf{Gym-Gazebo2}~\cite{lopezGymgazebo2ToolkitReinforcement2019s} has been developed since the beginning with the target of robotic applications. It interfaces with the Gazebo simulator, widely used in robotics. The environments are very easy to create thanks to the SDF format. Gazebo is a familiar tool to roboticists and laboratories can reuse many existing resources to interface their robotic platforms with the RL framework. The most significant drawback of the software architecture is the socket-based communication between the Python code of the environment and the simulator, that makes the simulation not fully reproducible.

\textbf{OpenAI ROS}\footnote{\url{http://wiki.ros.org/openai_ros}} provides RL environments for ROS robots running in the Gazebo simulator. Beyond sharing the same Gazebo drawbacks described for gym-gazebo2, Open Robotics did not yet implement the network segmentation that enables parallel simulations. Since the communication with the robot is based on the ROS middleware, this project in theory might support the application on real robots. Though, there is no suitable environment nor any documentation on how to execute or improve a policy on real robotic systems.

\textbf{Nvidia Isaac Gym}\footnote{\url{https://www.nvidia.com/en-us/deep-learning-ai/industries/robotics}} is the RL component of Isaac, the new Nvidia toolbox for AI applications in robotics. Simulations can be executed in either their PhysX or Flex engines and they provide state-of-the-art photorealistic rendering. Contrarily to Nvidia Isaac, which is already available, its Gym component that provides environments compatible with OpenAI Gym has not yet been released. Table~\ref{tab:comparison} was filled from the declared specifications and the previously presented work~\cite{liangGPUAcceleratedRoboticSimulation2018s} that will be integrated within the new framework. Isaac is one of the most promising projects that will provide a unified framework for robotics and AI, but unfortunately its closed source nature might limit the possibility of extending and customizing it.

\textbf{Unity ML Agent}~\cite{julianiUnityGeneralPlatform2018s} is another novel and promising toolkit for creating environments based on the Unity platform. It supports Nvidia PhysX out-of-the-box and plugins exist for Bullet and Mujoco. Being based on a gaming engine, rendering is very photorealistic. Despite agent code and physics engine residing on different processes, the selected gRPC communication protocol in its synchronous variant ensures determinism. However, custom actions and observations require the manual creation of the data serialization between client and server.

\textbf{Gibson}~\cite{xiaGibsonEnvRealWorld2018s} is another recent framework for active agents with main focus on real-world perception. Its rendering capabilities are highly photorealistic and they can be considered state-of-the-art. Its main target is simulation, and the physics runs in PyBullet.

\textbf{RaiSim}~\cite{hwangboPerContactIterationMethod2018s} is a recently released simulator specific for robotics. Its main advantage is an efficient contact solver that greatly speeds up the simulation. Due to its very recent release, there are not many examples available. As other frameworks, its closed-source nature might limit applications.

\textbf{PyRoboLearn}~\cite{delhaissePyRoboLearnPythonFramework2019s} is another framework containing both robotic environments and RL algorithms. It focuses on modularity and flexibility to promote code reuse. It currently supports only PyBullet, though it already has a physics engine abstraction layer in Python that will simplify adding other back-ends. A current limitation is the missing support to transfer code from simulation to real robots.

\begin{figure}
    \centering
    \resizebox{0.75\columnwidth}{!}{
        \input{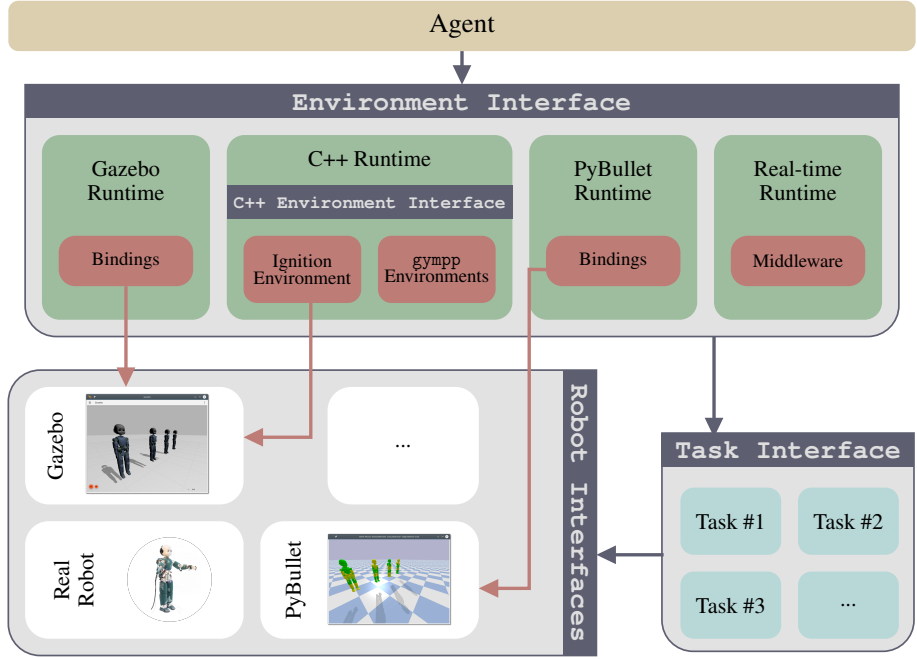}
    }
    \caption{Gym-Ignition software architecture. Dark rectangles represent the abstraction layers of the framework.}
    \label{fig:architecture}
\end{figure}

\begin{table}
    \footnotesize
    \center
    \caption{Comparison of frameworks that provide robotic environments compatible with OpenAI Gym.}
    \label{tab:comparison}
    \newcommand{\x}{\ensuremath{\times}}
    \newcommand{\ck}{\checkmark}
    \newcolumntype{Y}{>{\centering\arraybackslash}X}
\centerline{
    \begin{tabularx}{1.15\textwidth}{lYYYYYYYY}
        \toprule
        Software & Reproducible & Multiple Physics Engines & Photorealistic Rendering & Accelerated & Parallel & Real-Time Compatible & Modular & Open Source \\
        \midrule \rowcolor{black!20}
        OpenAI Robotic Environments &     &     &        & \ck & \ck &     & \ck    & $\sim$ \\
        Gym-Gazebo2                 &     & \ck &        & \ck & \ck &     &        & \ck    \\ \rowcolor{black!20}
        openai ros                  &     & \ck &        & \ck &     &     & \ck    & \ck    \\
        Bullet3 Environments        & \ck & \ck &        & \ck & \ck & \ck & $\sim$ & \ck    \\ \rowcolor{black!20}
        Nvidia Isaac Gym            & \ck & \ck & \ck    & \ck & \ck & \ck &        &        \\
        Unity ML-Agents             & \ck & \ck & \ck    & \ck & \ck &     &        & $\sim$ \\ \rowcolor{black!20}
        Gibson                      & \ck &     & \ck    & \ck & \ck &     & \ck    & \ck    \\
        RaiSim                      & \ck &     & \ck    & \ck & \ck &     &        &        \\ \rowcolor{black!20}
        PyRoboLearn            & \ck & \ck &        & \ck & \ck &     & \ck    & \ck    \\
        Gym-Ignition                & \ck & \ck & $\sim$ & \ck & \ck & \ck & \ck    & \ck    \\
        \bottomrule
    \end{tabularx}
    }
\end{table}

\section{The Proposed Framework: Gym-Ignition}

Gym-Ignition is designed for creating reproducible robotic environments with ease. Its architecture exploits widely interfaces and polymorphism, allowing to transparently change implementations while maintaining software abstraction. This design enables to develop tasks that can be executed on different robotic platforms both in simulation and in real-time.
Gym-Ignition components are described as follows:

\begin{itemize}
    \item \textbf{Environment:} The environment is a standard interface compliant with OpenAI's \texttt{Gym.Env}. Through the environment interface, the agent can reset the environment, set an action, and gather the observation and the associated scalar reward. Actions and observations are samples belonging to a specific space. 
    \item \textbf{Task:} The task is the interface that defines the logic about how to process the action received from the agent and to create the observation sample. It also calculates the reward and evaluates if the simulation reached its terminal state. In learning problems where the state is only partially observable, the task has typically access to the complete state, exposing to the agent only what necessary.
    \item \textbf{Robot:} The robot abstraction allows reading data from a robot and setting the references to be actuated. It is composed of a set of interfaces that permit to develop tasks agnostic on the robot. In fact, from the point of view of the task, the same method can gather data from either the simulated or real robot, depending on the active implementation. It also allows interfacing transparently with real robots that use different middlewares. The exposed \apis{} include joints, links, sensors, contacts, and robot base information.
    \item \textbf{Runtime:} Runtimes are the objects that expose the Environment interface to the agent. Runtimes are OpenAI Gym Wrappers that implement the final \texttt{Gym.Env} interface, and particularly its \texttt{step} method. They wrap the task and allow performing its execution in different simulators or in a real-time setting. 
\end{itemize}

As an example, let's consider a canonical toy problem in reinforcement learning: the cart-pole. In this case, the \emph{robot} is either a simulated or real cart-pole. Possible \emph{tasks} operating transparently on the cart-pole robot are pole balancing and swing-up. Finally, the task is wrapped by the selected \emph{runtime} and can be either simulated or executed in real-time.

Gym-Ignition is a project that aims to connect reinforcement learning libraries containing algorithms to common tools used in robotic research and industry. These two domains are historically related to Python and \cpp{} languages, respectively. Gym-Ignition allows implementing environments in Python and, experimentally, in \cpp.

Gym-Ignition implements most of the features we identified for frameworks that provide environments for robotics.
\textbf{Reproducibility.} In order to avoid socket-based client-server architectures, we designed the software architecture to enable running the simulation in the same process of the agent. Ignition Gazebo can be used as a library and we developed helper classes and SWIG~\cite{beazleySWIGEasyUse1996s} bindings to simplify its configuration and execution.
\textbf{Modularity.} The introduction of the Task, Robot, and Runtime abstractions provide the required flexibility to promote code reuse during environments development. Furthermore, this modularity can help minimizing system integration effort when changing the domain, particularly useful for sim-to-real applications.
\textbf{Real-time Compatibility.} The modular architecture allows wrapping the same task object that runs in a simulated runtime and execute it on the real robot without any change. The benefit is twofold. First, running a trained policy on the real robot would not require any code refactoring. Second, it would allow continuing the training process on the real robot since the switch is transparent from the point of view of the agent.
\textbf{Parallel Simulation.} Environment instances are independent to each other by design. The choice of not relying on client-server architectures greatly simplifies the development of parallel environments. Furthermore, in the case of distributed architectures, deployment and execution would be easier to setup since the simulator and the experience sampler can be embedded in a single component.
\textbf{Accelerated Simulation.} Ignition Gazebo supports tuning the real-time factor of its execution. Therefore, depending on the system resources and the scene complexity, simulations can run faster than real-time.
\textbf{Headless Simulation.} By default simulations run without any graphical interface. Though, visualizing the environment is helpful to design and debug the simulation, and the OpenAI Gym interface already offers a method to render the environment. When called, Gym-Ignition opens in a separated process a Gazebo GUI connected to the world of the simulated environment. The communication between the environment and the GUI happens with a client-server connection, but in this case it does not affect the reproducibility of the simulation being only uni-directional.
\textbf{Multiple Physics Engines.} The support of changing and adding physics engines is one of the main features of Gym-Ignition. A new engine can added in two ways: either in \cpp{} by implementing the physics interfaces of Ignition Robotics, or in Python by implementing a new Gym-Ignition Runtime and a new implementation of the Robot interfaces. We currently support DART~\cite{leeDARTDynamicAnimation2018s} through Gazebo, and PyBullet through our interfaces. 
\textbf{Photorealistic Rendering.} Gym-Ignition does not yet provide a complete support of photorealism since the default renderer is OGRE\footnote{\url{http://wiki.ogre3d.org}}. At the time of writing, Ignition Gazebo provides only a partial support of the Nvidia OptiX Engine~\cite{parkerOptiXGeneralPurpose2010s}.

Figure~\ref{fig:architecture} shows a diagram that represents the architecture of Gym-Ignition. It can be seen that the Environment Interface is the only Agent's entry-point to the decision-making task. Environments returned to the Agent through the OpenAI Gym factory are Runtime instances. We currently support Gazebo, PyBullet, Real-time, and \cpp{} runtimes. The red boxes show the underlying runtime components. The Gazebo and PyBullet runtimes communicate with the simulator through Python bindings. The Real-time runtime only needs to enforce a real-time execution of the Agent, and it communicates with the middleware of the robot platform to get the clock. Tasks contain the logic of the reinforcement learning problem and operate on a robot accessing its resources through the Robot Interfaces. This abstraction makes tasks robot-agnostic. Gym-Ignition provides implementation of the Robot Interfaces for all the supported Runtimes.

Robotics generally requires real-time constraints that are often difficult to enforce in Python-based applications. The \cpp{} Runtime, despite being still experimental, exists to meet this requirement. We developed \emph{gympp}, a \cpp{} port of the OpenAI Gym framework that maintains, where possible, the same structure and \apis. Gympp Environments can be either called as standalone executables or wrapped and exposed to Python. This approach enables developing intensive or real-time sensitive tasks in pure \cpp{}. As an example, we implemented a \cpp{} Gazebo Environment that, using the same library exposed to Python, can interface with Ignition Gazebo.

\section{Discussion and Conclusion}

This work presents Gym-Ignition, a novel framework to create reproducible robotic environments for reinforcement learning applications.
The main aim of the project is to narrow the gap between reinforcement learning and robotic research, allowing roboticists to create simulated environments with familiar tools. We believe that the quality and difficulty of the environments provided to the community (that define the decision-making problems to solve) are related to the scientific advances of this domain. We would also like to push the research outside the simulation realm, an extremely delicate step in the field of robotics. We hope that Gym-Ignition can motivate researchers on these two important directions.

Gym-Ignition enables the development of robotic environments with flexibility. Environments can be created either Python or \cpp{} and they can target either simulated or real robotic platforms. Thanks to Ignition Gazebo, physics engines can be switched with ease. The framework supports most of the simulation properties that enable to scale up the computations, such as accelerated, parallel, and distributed execution.

Gym-Ignition suffers from few limitations given its development status.
Despite the simulation can contain multiple robots in the same time, even interacting with each other, tasks can only operate on a single robot.
From the perception point of view, Ignition Gazebo already supports most of the common sensors typically mounted on robots, such as IMUs, lidars, and cameras. 
Few more sensors like force-torque will be implemented in future releases. Though, Gym-Ignition does not yet support gathering data from sensors. The partial support of photorealistic rendering engines does not make Gym-Ignition the best framework for visuomotor control. Though, the situation might improve as soon as the support of Nvidia OptiX will be finalized.
At last, given the early stage of development, the documentation of many components is either missing or improvable.

The ongoing and future activities on Gym-Ignition will introduce whole-body controllers for floating-base robots that can be used on all the supported physics engines.
We also plan to provide exhaustive domain randomization capabilities exposing to Python the physics engine parameters, sensors properties, kinematic and dynamic robot parameters. We are interested in extending the number of supported physics engines in Ignition Gazebo. In this regard, to push the randomization even further, we would like to include the entire physics engine in the domain.
The capability to insert models during runtime and the possibility to associate each of them to an independent task might represent a valid design for multi-agent RL. Despite the framework might be already capable of such simulations, we have not yet explored this applicability.
Lastly, we want to explore the integration with Ignition Fuel\footnote{https://app.ignitionrobotics.org/dashboard}, the new database maintained by Open Robotics containing worlds and 3D models of common objects and robots ready to be used. 
This integration can ease the creation of unstructured scenarios where robots operate and interact.

\section*{Acknowledgments}

This project has received funding from the European Union’s Horizon 2020 research and innovation programme under grant agreement No. 731540 (An.Dy).\newline
The content of this publication is the sole responsibility of the authors. The European Commission or its services cannot be held responsible for any use that may be made of the information it contains.

\bibliographystyle{vancouver}
\bibliography{references}

\end{document}